\documentclass[conference]{IEEEtran}
\IEEEoverridecommandlockouts
\usepackage{cite}
\usepackage{amsmath,amssymb,amsfonts}
\usepackage{algorithmic}
\usepackage{graphicx}
\usepackage{textcomp}
\usepackage{xcolor}

\usepackage{wrapfig}
\usepackage{tikz}
\usepackage{booktabs}
\usepackage{float}
\usepackage{subcaption}

\def\BibTeX{{\rm B\kern-.05em{\sc i\kern-.025em b}\kern-.08em
    T\kern-.1667em\lower.7ex\hbox{E}\kern-.125emX}}
\begin{document}



\title{Exposition and Interpretation of the Topology of Neural Networks \\
}

\author{\IEEEauthorblockN{Rickard Br\"uel Gabrielsson}
\IEEEauthorblockA{\textit{Department of Computer Science} \\
\textit{Stanford University}\\
Stanford, USA \\
rbg@cs.stanford.edu}
\and
\IEEEauthorblockN{Gunnar Carlsson}
\IEEEauthorblockA{\textit{Department of Mathematics} \\
\textit{Stanford University}\\
Stanford, USA \\
carlsson@stanford.edu}
}


\maketitle



\begin{abstract}
Convolutional neural networks (CNN's) are powerful and widely used tools. However, their interpretability is far from ideal. One such shortcoming is the difficulty of deducing a network's ability to generalize to unseen data. We use topological data
analysis to show that the information encoded in the weights of a CNN can be organized in terms of a topological data model and demonstrate how such information can be interpreted and utilized. We show that the weights of convolutional layers at depths from 1 through 13 learn simple global structures. We also demonstrate the change of the simple structures over the course of training. In particular, we define and analyze the spaces of spatial filters of convolutional layers and show the recurrence, among all networks, depths, and during training, of a simple circle consisting of rotating edges, as well as a less recurring unanticipated complex circle that combines lines, edges, and non-linear patterns. We also demonstrate that topological structure correlates with a network's ability to generalize to unseen data and that topological information can be used to improve a network's performance. We train over a thousand CNN's on MNIST, CIFAR-10, SVHN, and ImageNet.
\end{abstract}

\section{Introduction}

\begin{figure}[h]
\begin{center}
   \includegraphics[width=0.67\linewidth]{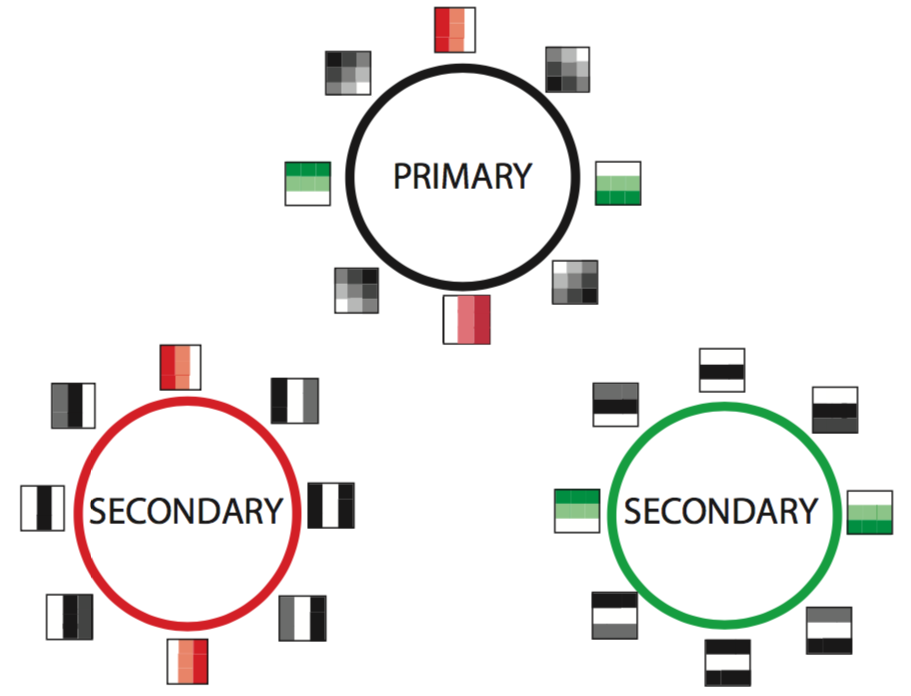}
\end{center}
	\vspace{-0.25cm}
   \caption{Primary and Secondary Circles.}
\label{fig:circles}
\vspace{-0.2cm}
\end{figure}

The problem of understanding how convolutional neural nets (CNN's) work and learn is one of the fundamental problems in machine learning. A related problem is CNN's tendency to overfit and be vulnerable to the so-called adversarial behavior, where by making tiny imperceptible changes to the input networks can be made to fail. In the context of neural networks, it is important to study both the weights and the activations, as these roughly constitute the "coefficients" and the outputs in the computational model.  To date, work in this area [1, 17, 18, 19, 20] has involved direct human inspection of features constructed in the network and has produced very interesting qualitative results. The first  goal of the present paper is to demonstrate that data sets constructed out of the weights are organized in simple {\em topological models}, which are strongly reminiscent of the results obtained in the topological analysis of data sets of local patches in natural images [2]. Such topological models yield insight by  effectively summarizing the global structure of the spaces of weight configurations, and permit the exploration of density in the data set. The key point here is that the study of the function of neural nets is a problem in {\em data analysis}, since the density of particular features is clearly relevant, and since we clearly find the presence of anomalous and spurious elements. It is important to model the most frequently occurring motifs in a simple and understandable way.  

\begin{table*}[h]
  \caption{\textit{M}(\textit{X}, \textit{Y}, \textit{Z}) CNN-architecture}
  \label{sample-table}
  \centering
  \begin{tabular}{llll}
    \toprule
    Conv Layer 1 & Conv Layer 2 & FC layer & Readout  \\
    \midrule
   3$ \times $3$ \times $\textit{X} filters  & 3$ \times $3$ \times $\textit{Y} filters & \textit{Z} nodes & 10 nodes \\
   ReLU & ReLU & ReLU & Softmax, Cross Entropy \\
   2$ \times $2 max-pooling & 2$ \times $2 max-pooling &  & Dropout 0.5, ADAM   \\
    \bottomrule
  \end{tabular}
  \label{table:M}
\end{table*}

\begin{table*}[h]
  \caption{\textit{C}(\textit{X}, \textit{Y}, \textit{Z}) CNN-architecture}
  \label{sample-table}
  \centering
  \begin{tabular}{llll}
    \toprule
    Conv Layer 1 & Conv Layer 2 & FC layer & Readout  \\
    \midrule
   3$ \times $3$ \times $\textit{X} filters  & 3$ \times $3$ \times $\textit{Y} filters & \textit{Z} nodes & 10 nodes \\
   ReLU  & ReLU & ReLU & Softmax \\
   3$ \times $3 max-pooling, stride: 2  & 2$ \times $2 max-pooling, stride: 1  &  & Cross Entropy    \\
   Local response normalization\footnotemark & Local response normalization\footnotemark[\value{footnote}] & & $L^2$ loss, SGD \\
    \bottomrule
  \end{tabular}
  \label{table:C}
\end{table*}

The topological models we work with are part of {\em topological data analysis} (TDA) [3, 9, 15, 16], which in addition to the construction of the models provide invariants of the shape of the data set (persistent homology), that confirm that the shape of the data is as expressed in the model.   We apply methods of TDA to data sets of {\em spatial filters} of the convolutional layers. In the $i$-th convolutional layer, an activation map is constructed by sliding a filter (a set of weights) along the spatial dimensions of all activation maps in the $(i-1)$-th layer. A filter thus has dimensions  $w \times h \times c$, where $w$ and $h$ are the width and height of the spatial receptive field of the filter while $c$ is the number of activation maps in the $(i-1)$-th layer. We define a {\em spatial filter} as one set of $w \times h$ weights with a fixed $c$-dimension. One single filter give $c$ spatial filters and a convolutional layer with $d$ number of activation maps give $d \times c$ spacial filters of dimension $w \times h$. 

We perform analyses of CNN's trained on the MNIST [5], CIFAR-10 [6] SVHN [22], and ImageNet [7] data sets. We find that in some cases, the models recapitulate the topological structures that occurred in [2], namely the {\em primary} and {\em secondary} circles (see Figure \ref{fig:circles}), but that in other  situations different  phenomena occur. The first part of this paper constitutes an exploratory analysis of the spatial filters described above. 

The second goal of this paper is to investigate the findings from the first part and provide an interpretation. Noticing that deeper networks with better generalizing abilities learn stronger topological structures, we estimate that topological structure is indicative of a network's ability to generalize to unseen data. If the topological structure learned by a network is a signature of its hypothesis about its task, then, in the spirit of Occam's razor, it is unlikely that a network learns a simple and strong topological structure that would only apply to its specific dataset at hand and not generalize to related data.

Indeed, we demonstrate that topological structure is indicative of a network's ability to generalize between the MNIST and SVHN datasets. Both datasets consists of images, MNIST consists of handwritten digits while SVHN is a more diverse dataset consisting of images of numbers for the addresses of houses. We show how one can improve the performance of a network trained on MNIST and evaluated on SVHN. We confirm that a network trained on SVHN generalizes better when evaluated on MNIST than vice versa, and show how the topological structure of a network trained on SVHN is 'simpler' than one trained on MNIST as predicted by our hypothesis. We also show how this measure of topological simplicity correlates with a networks performance on a held-out test set, for both MNIST and SVHN. Lastly, we show how extending a network with information obtained from our topological study may increase a networks performance on held out test data and speed up the learning process.

\footnotetext{With depth radius of 4}

\section{Persistent Homology}

Within the domain of topology, {\em homology} refers to a collection of signatures that perform a sophisticated counting task for features, such as connected components, loops, spheres, etc. to obtain invariants of topological spaces. Its extension to point clouds is called {\em persistent homology}, which has been undergoing rapid development over the last 15 years. For each dimension $k$, the output of persistent homology is a {\em barcode}, i.e. an unordered collection of intervals on the real line, where a long bar indicates the presence of a feature that lives over a large range of values and is hence regarded as “real”, and short bars are often attributed to noise. The barcode is a multiscale summary analogous to the dendrograms that arise in hierarchical clustering. For example, a long bar in the 1-dimensional bar code reflects the presence of a loop in the data. These invariants have been used in many different situations. One such is the analysis of local image patches performed in [2], which was motivated by the idea of understanding the tuning of neurons in the primary visual cortex. One of the outcomes of that paper is illustrated above (Figure \ref{fig:circles}), where we see that the data (suitably thresholded by density) is organized around three circles, which overlap to a degree, and which reflect the tuning of neurons to edge and line detectors. The idea of this paper is to perform this same analysis in the context of neural nets rather than the visual pathway. 


\section{Mapper}

The topological modeling method (''Mapper", see  [10] for details) we use starts with one, two, or three real valued functions on the data, which we refer to as {\em lenses}, as well as with a metric on the data set.  By choosing overlapping coverings of the real line by intervals of the same length and overlap, we obtain coverings of $\Bbb{R}$, $\Bbb{R}^2$, or $\Bbb{R}^3$, which allow us to group the data into bins, one for each set in the cover.  We then perform a clustering step (single linkage clustering with a fixed heuristic for the choice of threshold, specified in [12]) based on the metric to generate a set of clusters.  Because the intervals overlap, it is possible for clusters attached to one bin to overlap with clusters attached to another bin, and we define a graph whose node set is the collection of clusters we have defined, and where there is an edge connecting a pair of clusters  if the two clusters share at least one data point.  The topological version of this construction is well known, and comes with guarantees concerning the degree to which the construction approximates the original space. Such guarantees are not yet available for Mapper, although work in this direction is being done [11].  

For the clustering step in the Mapper method we use the Variance Normalized Euclidean (VNE) metric. The VNE metric is a variant of standard Euclidean distance that first normalizes each column of the data set by dividing by its variance. For lenses we use PCA 1 and 2, which means that the point cloud is projected onto its two principal components before choosing overlapping coverings. Our results generalize to other lenses such as Ayasdi's Neighborhood Lens 1 and 2 [14] which capture more non-linear features of the data. However, since PCA lenses often gave the best-looking graphs and for sake of consistency and simplicity we only present results acquired by use of the PCA lenses. We use the implementation of Mapper found in the Ayasdi software [12]. In Ayasdi, {\em resolution} specifies the number of bins and {\em gain} determines the overlap as follows: $ \textrm{percent overlap}\ = 1 - (1/gain) $. We specify Mapper by notation $Mapper(resolution, gain)$. In addition, the color of the nodes is determined by the number of points that the corresponding cluster contains, with red being the largest and blue the lowest. This number is a rough proxy for density.

\section{Density Filtration}

To determine the core subset of a point cloud $X$ we perform a density filtration of the points based on a nearest neighbor estimate of the local density. For each $x\in X$ and $k>0$ we calculate its distance to its $k$-th nearest neighbor, this distance being inversely correlated to the density at $x$. Then we take the top $p, 0 < p \leq 1$, fraction of the densest points. We can thus denote a density filtration with parameters $k$ and $p$ applied to $X$ by $\rho(k,p,X)$.

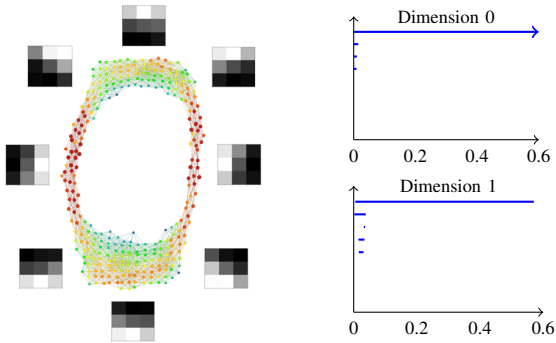
\begin{figure}[h]
\centering
\begin{minipage}{.45\textwidth}
  \centering
 \begin{minipage}{.53\textwidth}
  \includegraphics[height=4.6cm]{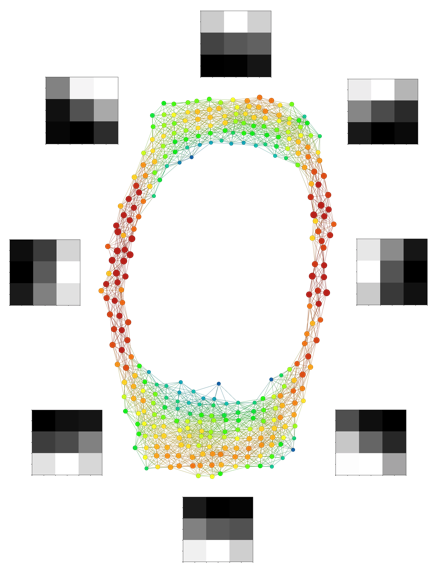} 	
 \end{minipage}
  \begin{minipage}{.40\textwidth}

\begin{tikzpicture}[scale = 0.41]

\draw[->] (0,0) -- (6,0) node[anchor=north] {};

\draw	(3,4) node{{\scriptsize Dimension 0}};

\draw	(0,0) node[anchor=north] {\scriptsize 0}
		(2,0) node[anchor=north] {\scriptsize 0.2}
		(4,0) node[anchor=north] {\scriptsize 0.4}
		(6,0) node[anchor=north] {\scriptsize 0.6};
\draw	(1,3.5) node{{}}
		(4,3.5) node{{}};

\draw[->] (0,0) -- (0,4) node[anchor=east] {};

\draw[->, thick, blue] (0,3.5) -- (6,3.5);
\draw[thick, blue] (0,3.1) -- (0.15,3.1);
\draw[thick, blue] (0,2.7) -- (0.10,2.7);
\draw[thick, blue] (0,2.3) -- (0.09,2.3);

\end{tikzpicture}

\begin{tikzpicture}[scale = 0.42]

\draw[->] (0,0) -- (6,0) node[anchor=north] {};

\draw	(3,4) node{{\scriptsize Dimension 1}};

\draw	(0,0) node[anchor=north] {\scriptsize 0}
		(2,0) node[anchor=north] {\scriptsize 0.2}
		(4,0) node[anchor=north] {\scriptsize 0.4}
		(6,0) node[anchor=north] {\scriptsize 0.6};
\draw	(1,3.5) node{{}}
		(4,3.5) node{{}};

\draw[->] (0,0) -- (0,4) node[anchor=east] {};

\draw[thick, blue] (0.05,3.5) -- (5.7,3.5);
\draw[thick, blue] (0.02,3.1) -- (0.38,3.1);
\draw[thick, blue] (0.32,2.7) -- (0.36,2.7);
\draw[thick, blue] (0.15,2.3) -- (0.34,2.3);
\draw[thick, blue] (0.16,1.9) -- (0.31,1.9);

\end{tikzpicture}

  \end{minipage}
  \caption{$Mapper(30,3)$ and barcodes of $ \rho(200,0.3, 100 \times M^{1}_{40K}(64,32,64))$}
  \label{fig:mnist_1st_2layer}
\end{minipage}%
\end{figure}


\section{Topological Analysis of Weight Spaces}

Our first experiments were conducted on networks trained on the MNIST [5], CIFAR-10 [6], and ImageNet [7] datasets. MNIST consists of gray scale images of digits, CIFAR-10 consists of natural color images of 10 classes including airplanes, cats, dogs, and ships, and ImageNet consists of natural color images of a wide variety of classes. CNN's have achieved high accuracy all these data sets, suggesting that CNN's are able to learn structures present among the images in the data sets.  

We specify the architecture of our CNN's as in Table \ref{table:M} and \ref{table:C}, where \textit{X, Y, Z} corresponds to the depth of the first convolutional layer, the depth of the second convolutional layer, and the number of nodes in the fully connected layer respectively. If any of \textit{X, Y}, or \textit{Z} is $0$ it means that that whole column or block is removed from the network. E.g. $M(64,32,64)$ is a network of type found in Table \ref{table:M} with a first-convolutional-layer-depth of 64, a second-convolutional-layer-depth of 32, followed by a fully connected layer with 64 nodes. For notational efficiency we use superscripts to specify the convolutional layer from which the spatial filters were extracted and subscripts to specify the number of batch iterations the network was trained on. Further, preceding this notation by '$N \times$' means that $N$ trained networks were used as the source of the spatial filters. Thus, with previously developed notation we can write $Mapper(30,3)$ of $ \rho(200,0.2, 100 \times M^{1}_{100K}(64,32,64))$ to denote Mapper with resolution 30 and gain 3 applied to a point cloud generated by a k-nearest-neighbor filtration with $k=200$, $p=0.2$ of the mean-centered and normalized $1$st convolutional layers' spatial filters of 100 networks of type $M(64,32,64)$ trained for 100,000 batch iterations. Throughout this work we treat each spatial filter of a convolutional layer as a point, i.e. each point is (\textit{width}$\times$\textit{height})-dimensional. We \textit{always} mean-center and normalize each point, which is done before any density filtration. In addition, the padding on the convolutional layers preserves spacial dimensionality and a batch size of 124 was used throughout the experiments. 

\subsection{MNIST}

MNIST was divided into 60,000 training examples and 10,000 test examples. We train 100 CNN's of type \textit{M}(64,32,64) (Table \ref{table:M}) for 40,000 batch iterations with a batch size of 128 to a test accuracy of about $ 99.0 \% $. These 100 trained CNN's give us $64 \times 100=6400$ 9-dimensional points (first layer spatial filters) which we mean-center and normalize. We then use k-nearest-neighbor density filtration with $k=200$ and $p=0.3$ to get $1920$ points. To this point cloud (equivalent to $ \rho(200,0.3, 100 \times M^{1}_{40K}(64,32,64))$) we apply Mapper ($resolution=30$, $gain=3$) with Variance Normalized Euclidean Norm and two PCA lenses. The resulting graph can be seen in Figure \ref{fig:mnist_1st_2layer}. We also put, next to the graph, the mean of adjacent points to represent the spatial filters at that position in the graph. Recall that color codes for the size of the collection represented by the nodes, increasing from blue to red.

\begin{figure}[h]
\centering
 \includegraphics[height=3.9cm]{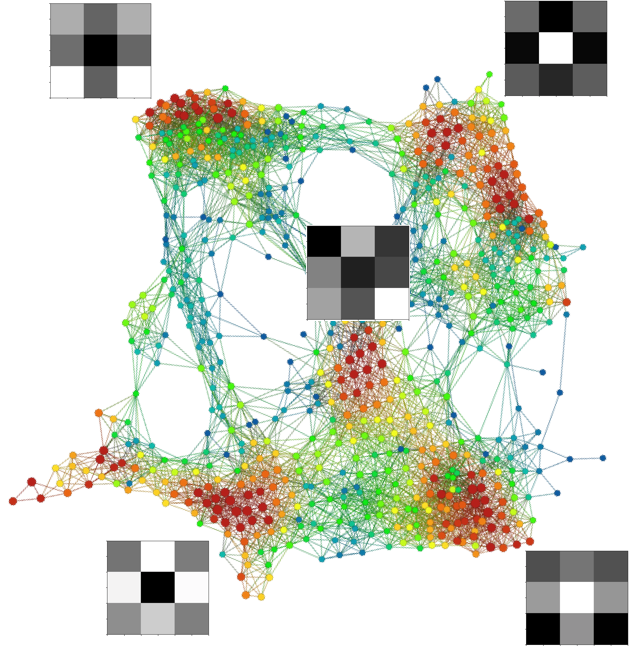}
\hspace{0.2cm}
 \includegraphics[height=3.9cm]{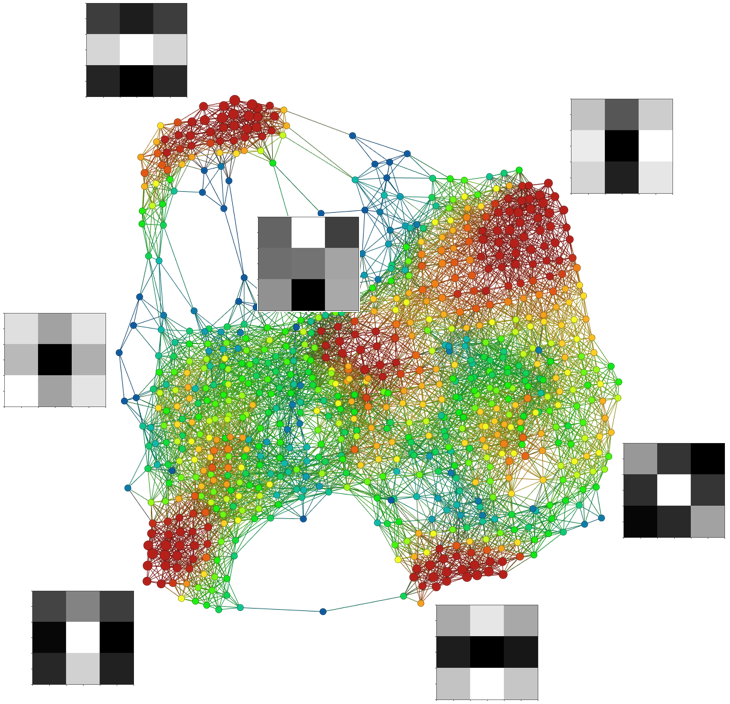}
\caption{$Mapper(30,3)$ of $ 48 \times C^{1}_{70K}(64,0,64)$, and $Mapper(30,3)$ of $ \rho(200,0.5, 100 \times C^{1}_{70K}(64,32,64))$}
\label{fig:cifargray_1st_2layer_mapper}
\end{figure}

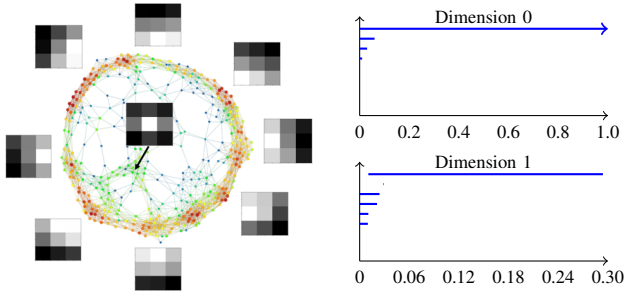
\begin{figure}[h]
\centering
\begin{subfigure}{0.23\textwidth}
\centering
        \includegraphics[height=3.9cm]{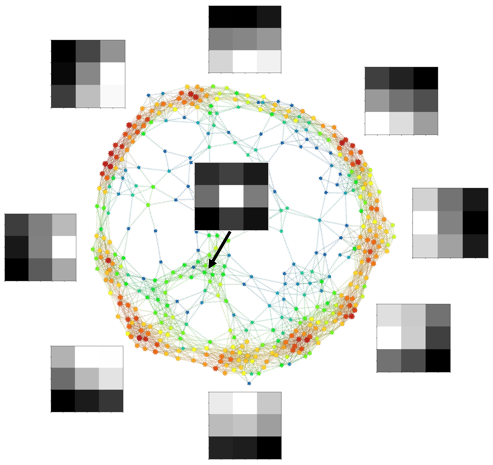} 
\end{subfigure}
\hspace{0.1cm}
\begin{subfigure}{0.20\textwidth}
\centering
\begin{tikzpicture}[scale = 0.33]
\draw[->] (0,0) -- (10,0) node[anchor=north] {};

\draw	(5,4) node{{\scriptsize Dimension 0}};

\draw	(0,0) node[anchor=north] {\scriptsize 0}
		(2,0) node[anchor=north] {\scriptsize 0.2}
		(4,0) node[anchor=north] {\scriptsize 0.4}
		(6,0) node[anchor=north] {\scriptsize 0.6}
		(8,0) node[anchor=north] {\scriptsize 0.8}
		(10,0) node[anchor=north] {\scriptsize 1.0};
\draw	(1,3.5) node{{}}
		(4,3.5) node{{}};

\draw[->] (0,0) -- (0,4) node[anchor=east] {};

\draw[->, thick, blue] (0,3.5) -- (10,3.5);
\draw[thick, blue] (0,3.1) -- (0.6,3.1);
\draw[thick, blue] (0,2.7) -- (0.3,2.7);
\draw[thick, blue] (0,2.3) -- (0.1,2.3);

\end{tikzpicture}
\begin{tikzpicture}[scale = 0.33]

\draw[->] (0,0) -- (10,0) node[anchor=north] {};

\draw	(5,4) node{{\scriptsize Dimension 1}};

\draw	(0,0) node[anchor=north] {\scriptsize 0}
		(2,0) node[anchor=north] {\scriptsize 0.06}
		(4,0) node[anchor=north] {\scriptsize 0.12}
		(6,0) node[anchor=north] {\scriptsize 0.18}
		(8,0) node[anchor=north] {\scriptsize 0.24}
		(10,0) node[anchor=north] {\scriptsize 0.30};
\draw	(1,3.5) node{{}}
		(4,3.5) node{{}};

\draw[->] (0,0) -- (0,4) node[anchor=east] {};

\draw[thick, blue] (0.35,3.5) -- (9.8,3.5);
\draw[thick, blue] (0.95,3.1) -- (0.97,3.1);
\draw[thick, blue] (0,2.7) -- (0.8,2.7);
\draw[thick, blue] (0,2.3) -- (0.7,2.3);
\draw[thick, blue] (0,1.9) -- (0.35,1.9);
\draw[thick, blue] (0,1.5) -- (0.33,1.5);
\end{tikzpicture}
\end{subfigure}
\caption{$Mapper(30,3)$ of $ \rho(75,0.37,C^{2}_{60K}(64,32,64))$, and barcodes of $ \rho(15,0.1, 100 \times C^{2}_{50K}(64,32,64))$}
  \label{fig:cifargray_2nd_2layer_barcode}
\end{figure}

From this graph we see how the learned spatial filters are well approximated by the primary circle (Figure \ref{fig:circles}). The circle is further supported by the corresponding barcodes (Figure \ref{fig:mnist_1st_2layer}), which show one persistent loop or circle and one persistent connected component. We obtain almost identical results as in Figure \ref{fig:mnist_1st_2layer} with $Mapper(30,3)$ and barcodes of $ \rho(200,0.3, 100 \times M^{1}_{40K}(64,0,64))$, i.e. only having one convolutional layer. The results were also robust to other network configurations; the primary circle was found in the first layer spatial filters of trained networks of types $M(64,8,512)$, $M(64,16,512)$, and $M(256,32,512) $.

For the same networks of type $M(64,32,64)$ used to generate Figure \ref{fig:mnist_1st_2layer} we also obtain $64 \times 32 \times 100 = 204800 $ 9-dimensional second layer spatial filters. After strong density filtration ($p=0.1$, $k=10$) we find a very weak primary circle: significantly weaker than that found in the first layer.

\subsection{CIFAR-10}

CIFAR-10 was divided into 50,000 training examples and 10,000 test examples. The input was preprocessed by taking a random $24 \times 24$ crop of the image, applying a random left-right flip, mean-centering, and normalizing. 

\begin{figure*}[h]
\centering
\begin{minipage}{1.0\textwidth}
\center
\includegraphics[height=5.1cm]{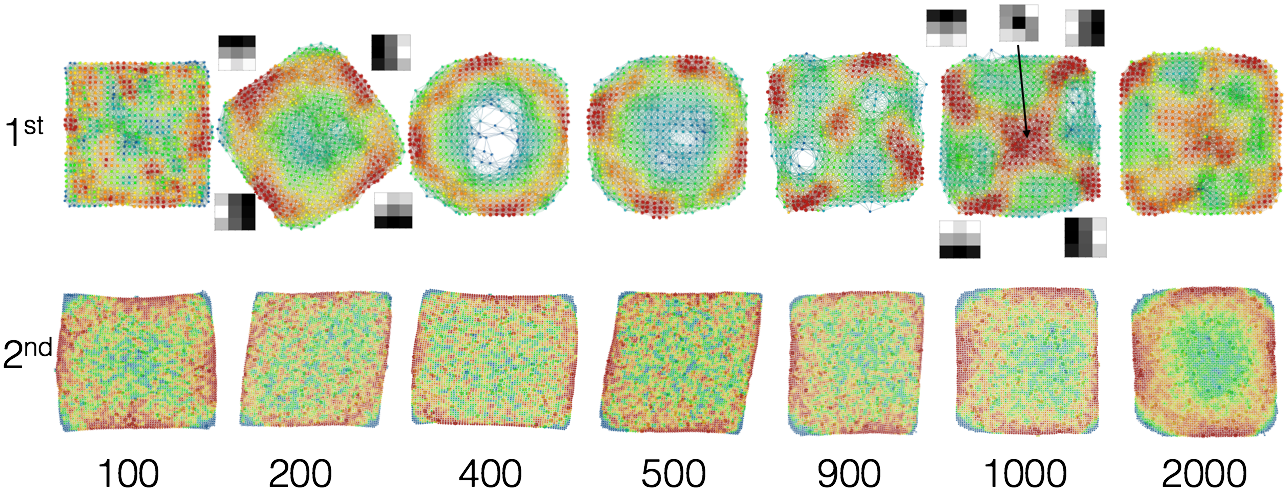} 
  \caption{$Mapper(30,3)$ of $ 100 \times C^{1}(64,32,64)$ and $Mapper(70,2)$ of $ \rho(15,0.5, 100 \times C^{2}(64,32,64))$ from 100-2000 batch iterations. Best viewed in color.}
  \label{fig:cifargray_learning}
\end{minipage}%
\end{figure*}

\begin{figure}[h]
\center
\includegraphics[height=4.4cm]{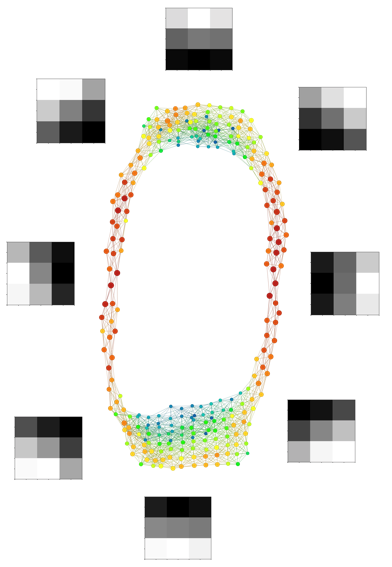}
\hspace{0.2cm}
\includegraphics[height=4.4cm]{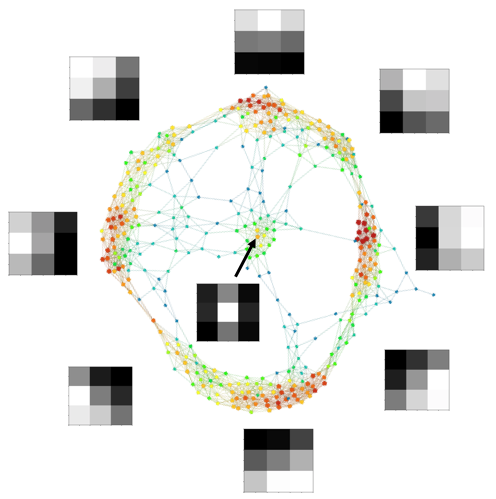} 
  \caption{$Mapper(30,3)$ of $ \rho(200, 0.14, 60 \times C^{1}_{100K}(64,32,64))$, and $Mapper(30,3)$ of $ \rho(10, 0.32, C^{2}_{50K}(64,32,64))$}
  \label{fig:cifar_1st_2layer_mapper}
\end{figure}

\begin{figure}[h]
  \centering	
	\begin{subfigure}{.23\textwidth}
        \centering
        \includegraphics[height=3.9cm]{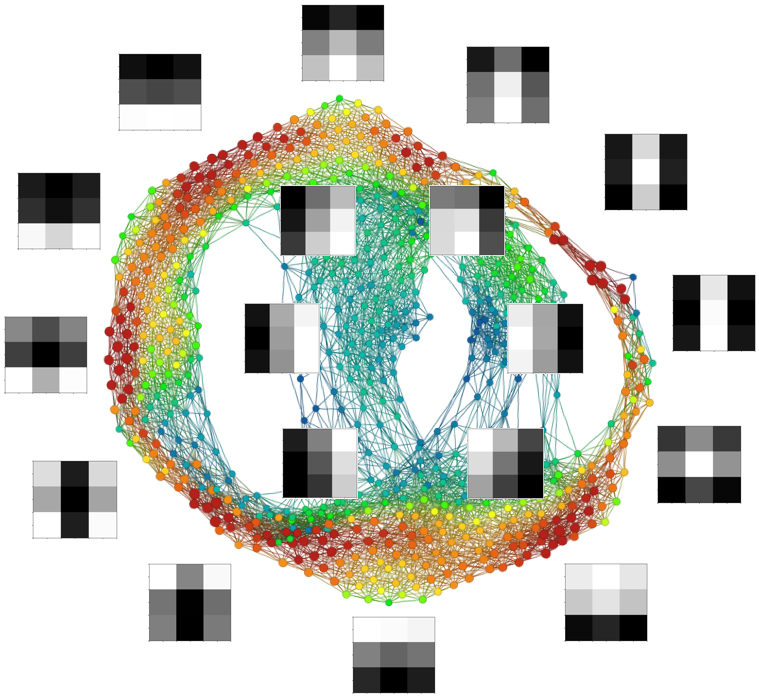} 
    \end{subfigure}
    \hspace{0.1cm}
    \begin{subfigure}{.20\textwidth}
\begin{tikzpicture}[scale = 0.33]

\draw[->] (0,0) -- (10,0) node[anchor=north] {};

\draw	(5,4) node{{\scriptsize Dimension 0}};

\draw	(0,0) node[anchor=north] {\scriptsize 0}
		(2,0) node[anchor=north] {\scriptsize 0.08}
		(4,0) node[anchor=north] {\scriptsize 0.16}
		(6,0) node[anchor=north] {\scriptsize 0.24}
		(8,0) node[anchor=north] {\scriptsize 0.32}
		(10,0) node[anchor=north] {\scriptsize 0.40};
\draw	(1,3.5) node{{}}
		(4,3.5) node{{}};

\draw[->] (0,0) -- (0,4) node[anchor=east] {};

\draw[->, thick, blue] (0,3.5) -- (10,3.5);
\draw[thick, blue] (0,3.1) -- (0.2,3.1);
\draw[thick, blue] (0,2.7) -- (0.18,2.7);
\draw[thick, blue] (0,2.3) -- (0.16,2.3);

\end{tikzpicture}

\begin{tikzpicture}[scale = 0.33]

\draw[->] (0,0) -- (10,0) node[anchor=north] {};

\draw	(5,4) node{{\scriptsize Dimension 1}};

\draw	(0,0) node[anchor=north] {\scriptsize 0}
		(2,0) node[anchor=north] {\scriptsize 0.08}
		(4,0) node[anchor=north] {\scriptsize 0.16}
		(6,0) node[anchor=north] {\scriptsize 0.24}
		(8,0) node[anchor=north] {\scriptsize 0.32}
		(10,0) node[anchor=north] {\scriptsize 0.40};
\draw	(1,3.5) node{{}}
		(4,3.5) node{{}};

\draw[->] (0,0) -- (0,4) node[anchor=east] {};

\draw[thick, blue] (0.5, 3.5) -- (9.3, 3.5);
\draw[thick, blue] (0.4, 3.1) -- (9.3, 3.1);
\draw[thick, blue] (0.6, 2.7) -- (9.2, 2.7);
\draw[thick, blue] (0.2, 2.3) -- (1.1, 2.3);
\draw[thick, blue] (0.8, 1.9) -- (0.9, 1.9);
\draw[thick, blue] (0.18, 1.5) -- (0.4, 1.5);
\draw[thick, blue] (0.18, 1.1) -- (0.35, 1.1);
\draw[thick, blue] (0.22, 0.7) -- (0.3, 0.7);

\end{tikzpicture}

\end{subfigure}
\caption{$Mapper(30,3)$ and barcodes of $ \rho(200, 0.32, 82 \times C^{*1}_{30K}(48,0,64))$. *: Without max-pooling}
\label{fig:cifar_1st_1layer}
\end{figure}
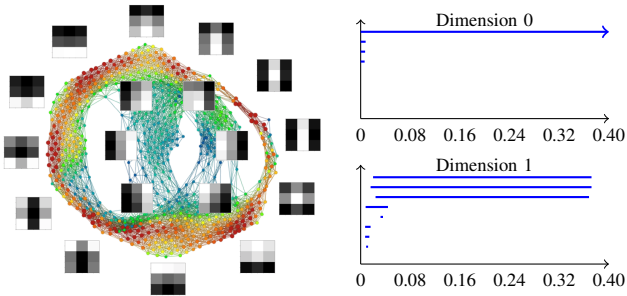   

\begin{figure}[h]
  \centering
  \centering	
	\begin{subfigure}{.23\textwidth}
        \centering
        \includegraphics[height=4.0cm]{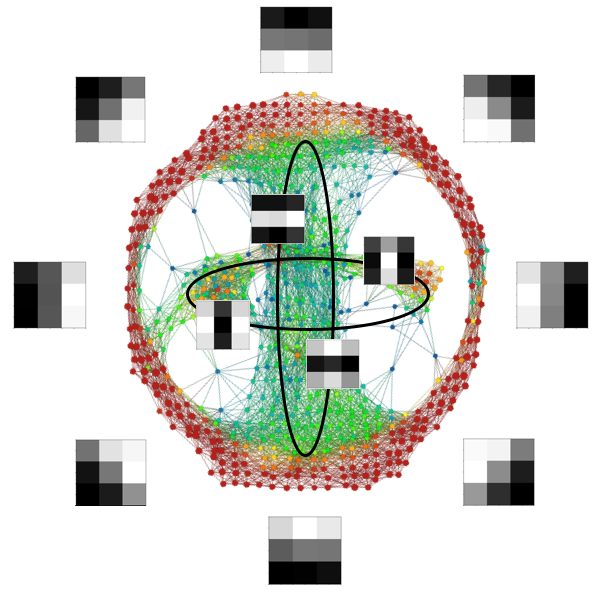} 
    \end{subfigure}
    \hspace{0.1cm}
    \begin{subfigure}{.20\textwidth}

\begin{tikzpicture}[scale = 0.33]

\draw[->] (0,0) -- (10,0) node[anchor=north] {};

\draw	(5,4) node{{\scriptsize Dimension 0}};

\draw	(0,0) node[anchor=north] {\scriptsize 0}
		(2,0) node[anchor=north] {\scriptsize 0.06}
		(4,0) node[anchor=north] {\scriptsize 0.12}
		(6,0) node[anchor=north] {\scriptsize 0.18}
		(8,0) node[anchor=north] {\scriptsize 0.24}
		(10,0) node[anchor=north] {\scriptsize 0.30};
		
\draw	(1,3.5) node{{}}
		(4,3.5) node{{}};

\draw[->] (0,0) -- (0,4) node[anchor=east] {};

\draw[->, thick, blue] (0,3.5) -- (10,3.5);
\draw[thick, blue] (0,3.1) -- (3,3.1);
\draw[thick, blue] (0,2.7) -- (0.7,2.7);
\draw[thick, blue] (0,2.3) -- (0.4,2.3);

\end{tikzpicture}

\begin{tikzpicture}[scale = 0.33]

\draw[->] (0,0) -- (10,0) node[anchor=north] {};

\draw	(5,4) node{{\scriptsize Dimension 1}};

\draw	(0,0) node[anchor=north] {\scriptsize 0}
		(2,0) node[anchor=north] {\scriptsize 0.06}
		(4,0) node[anchor=north] {\scriptsize 0.12}
		(6,0) node[anchor=north] {\scriptsize 0.18}
		(8,0) node[anchor=north] {\scriptsize 0.24}
		(10,0) node[anchor=north] {\scriptsize 0.30};
		
\draw	(1,3.5) node{{}}
		(4,3.5) node{{}};

\draw[->] (0,0) -- (0,4) node[anchor=east] {};

\draw[thick, blue] (0.5, 3.5) -- (8.6, 3.5);
\draw[thick, blue] (0.4, 3.1) -- (8.4, 3.1);
\draw[thick, blue] (0.1, 2.7) -- (8.4, 2.7);
\draw[thick, blue] (4.7, 2.3) -- (8.3, 2.3);
\draw[thick, blue] (4.6, 1.9) -- (6.7, 1.9);
\draw[thick, blue] (0.1, 1.5) -- (1.0, 1.5);
\draw[thick, blue] (0.7, 1.1) -- (0.8, 1.1);
\draw[thick, blue] (0.1, 0.7) -- (0.8, 0.7);
\draw[thick, blue] (0.2, 0.3) -- (0.7, 0.3);

\end{tikzpicture}

\end{subfigure}  
\caption{$Mapper(30,3)$ and barcodes of $ \rho(100, 0.35, 100 \times C^{1}_{100K}(64,32,64))$. }
 \label{fig:cifar_1st_2layer_c3}
\end{figure}
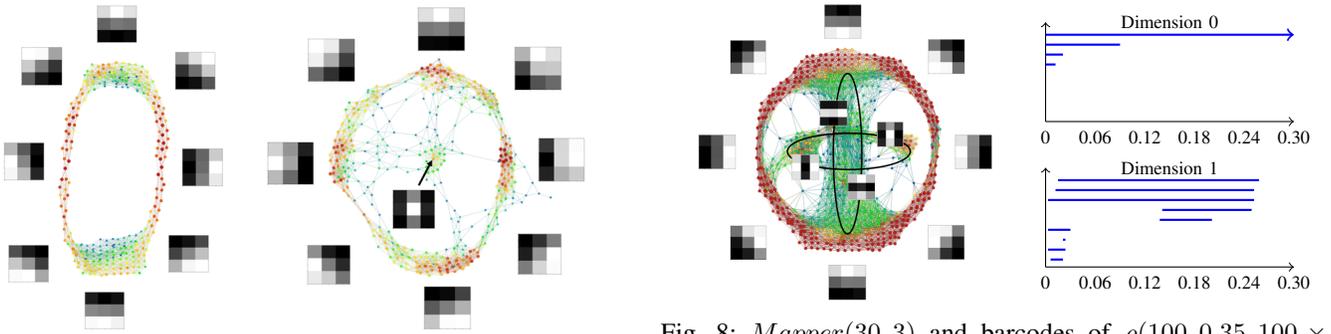

\begin{figure*}[h]
\centering
\begin{minipage}{1.0\textwidth}
  \centering
 \includegraphics[width=0.83\textwidth]{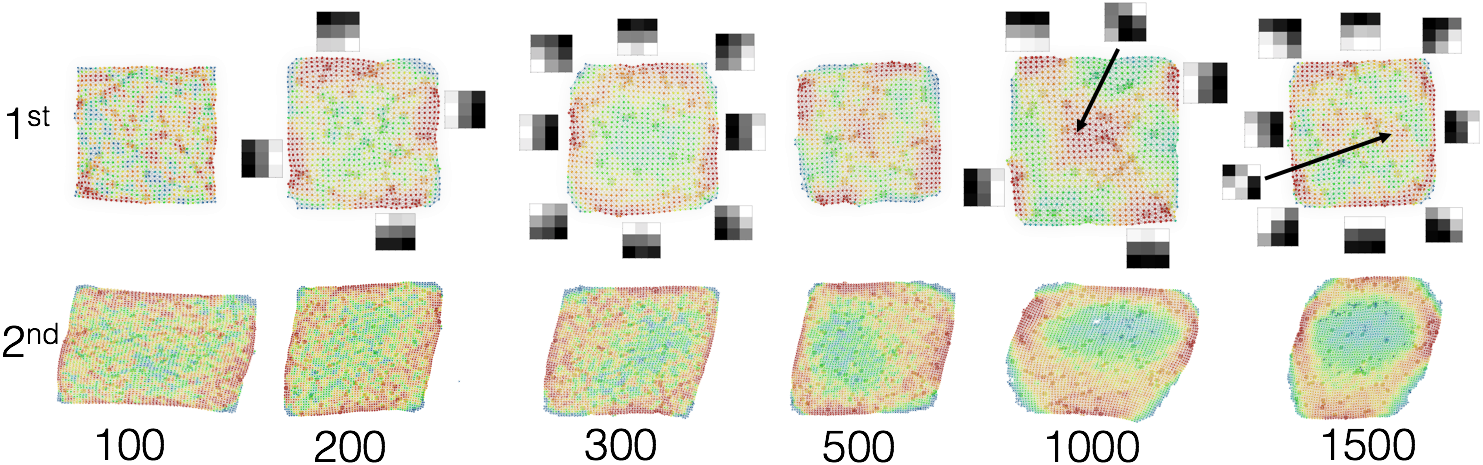} 
 \caption{$Mapper(33,2)$ of $ 60 \times C^{1}(64,32,64)$ and $Mapper(60,2)$ of $\rho(100, 0.3, 60 \times C^{2}(64,32,64))$ from 100-1500 batch iterations. Best viewed in color.}
  \label{fig:cifar_learning}
\end{minipage}%
\end{figure*}

\subsubsection{Grayscaled}

The input was grayscaled using the weights $(0.2989, 0.5870, 0.1140)$ for red, green, and blue respectively. We train 100 CNN's of configuration $C(64,32,64)$ for 70,000 batch iterations (test accuracy of about 77.0\%) to obtain 6,400 first-layer spatial filters and 204,800 second-layer spatial filters. The result of $(p=0.5, k=200$) density filtration and Mapper on the first-layer spatial filters can be seen in Figure \ref{fig:cifargray_1st_2layer_mapper}. We also train 48 CNN's of configuration $ C(64,0,64) $ for 70,000 batch iterations (test accuracy of about 69.2\%) to obtain 3,072 first-layer spatial filters; the result of Mapper on these first-layer filters can also be see in Figure \ref{fig:cifargray_1st_2layer_mapper}. Notice that that in both cases we find five cluster structures but that the clusters differ between the two cases. In the latter we find clusters around horizontal and vertical lines while this is not the case in the former. In neither of the 'well-trained' networks were we able to find a significant primary circle.

In Figure \ref{fig:cifargray_2nd_2layer_barcode} we show the barcodes of the 204,800 second-layer spatial filters from the 100 CNN's of configuration $C(64,32,64)$ trained for 50,000 batch iterations (test accuracy of about 76.2\%) and with density filtration $p=0.1, k=15$. In the same Figure we also show Mapper applied to the 2,048 second-layer spatial filters of a single CNN of configuration $C(64,32,64)$ trained for 60,000 batch iterations (test accuracy of about 77.1 \%), and with density filtration $p=0.37, k=75$. Note that even though we needed more networks to get the clear barcodes in Figure \ref{fig:cifargray_2nd_2layer_barcode} showing the circle, the Mapper output in the same Figure demonstrates that the primary circle (with some other weaker structures) appears in the training of a single network.

Next we look at the spatial filters of the first and second convolutional layers of 100 CNN's of configuration $C(64,32,64)$ at batch iterations 100 to 2000. In Figure \ref{fig:cifargray_learning} we see Mapper applied to both these point clouds. The vertical axis specifies the index of the convolutional layer (1st or 2nd) and the horizontal axis specifies the number of batch iterations. For the 2nd layer spatial filters a density filtration of $p=0.5, k=15$ was applied, while no density filtration was applied to the first layer. We find that in the first layer the primary circle reveals itself at 400 batch iterations, breaks apart at 500 batch iterations, and then starts to reappear in the second layer at 2000 batch iterations. Note that the four edges in the first layer shown at 200 and 1000 iterations appear relatively stable over many batch iterations.

\subsubsection{Color}

We train 60 CNN's of configuration $C(64,32,64)$ for 100,000 batch iterations (test accuracy of about 81.2\%). This gives us 11,520 first-layer spatial filters and 204,800 second-layer spatial filters. In Figure \ref{fig:cifar_1st_2layer_mapper} we show Mapper applied to the 11,520 first layer spatial filters at 100,000 batch iterations and density filtration $p=0.14, k=200$. In the same Figure we also show Mapper applied to the 2,048 second layer spatial filters of a single network at 50,000 batch iterations (test accuracy of about 79.9\%) and density filtration $p=0.32, k=10$. 

We also compute the barcodes of the point cloud of the first-layer spatial filters and find an equally persistent circle and connected component as in the barcodes of Figure \ref{fig:mnist_1st_2layer}. We also compute the barcodes of all the 204,800 second-layer spatial filters at 100,000 batch iterations and density filtration $p=0.1, k=15$ and find similar support for the circle as found in the gray scaled case of Figure \ref{fig:cifargray_2nd_2layer_barcode}. In addition, we look at the first layer spatial filters for each input channel, i.e. red, green, and blue, independently and find the primary circle in each one. 

Next we train 82 CNN's of configuration $C(48,0,64)$ but \textit{without} max-pooling and find among the 11,808 first layer spacial filters at 30,000 batch iterations (test accuracy of about 71.8\%) and filtration $p=0.32, k=200$ the two-circle model showed in Figure \ref{fig:cifar_1st_1layer}. We see that the circles intersect at two points and that one of the circles (the weaker) is the primary circle while the other (the stronger) is a strange circle we have not seen before. Two circles intersecting at two points have three loops and one connected component, which can be seen among the barcodes in Figure \ref{fig:cifar_1st_1layer}. 

A closer examination of the 11,520 first-layer spatial filters of the configuration $C(64,32,64)$, trained for 100,000 batch iterations at filtration $p=0.35, k = 100$, shows that the three circle model found in the image patch data [2] appears. The barcodes and Mapper applied to this point cloud can be seen in Figure \ref{fig:cifar_1st_2layer_c3}. Note the stronger outer primary circle and the two weaker secondary circles; each of the secondary circles intersect the primary circle twice but they do not intersect each other. 

We look at the spatial filters of the first and second convolutional layers of 60 CNN's of configuration $C(64,32,64)$ at batch iterations 100 to 1500. In Figure \ref{fig:cifar_learning} we see Mapper applied to both these point clouds. The vertical axis specifies the index of the convolutional layer (1st or 2nd) and the horizontal axis specifies the number of batch iterations. Note, in the first layer, that the primary circle appears at 300 batch iterations, breaks apart at 500 iterations, and then reappears at 1500 batch iterations with some inner secondary structures. The primary circle appears in the second convolutional layer at 1000 batch iterations. 

\begin{figure*}[h]
\begin{center}
\includegraphics[width=0.83\linewidth]{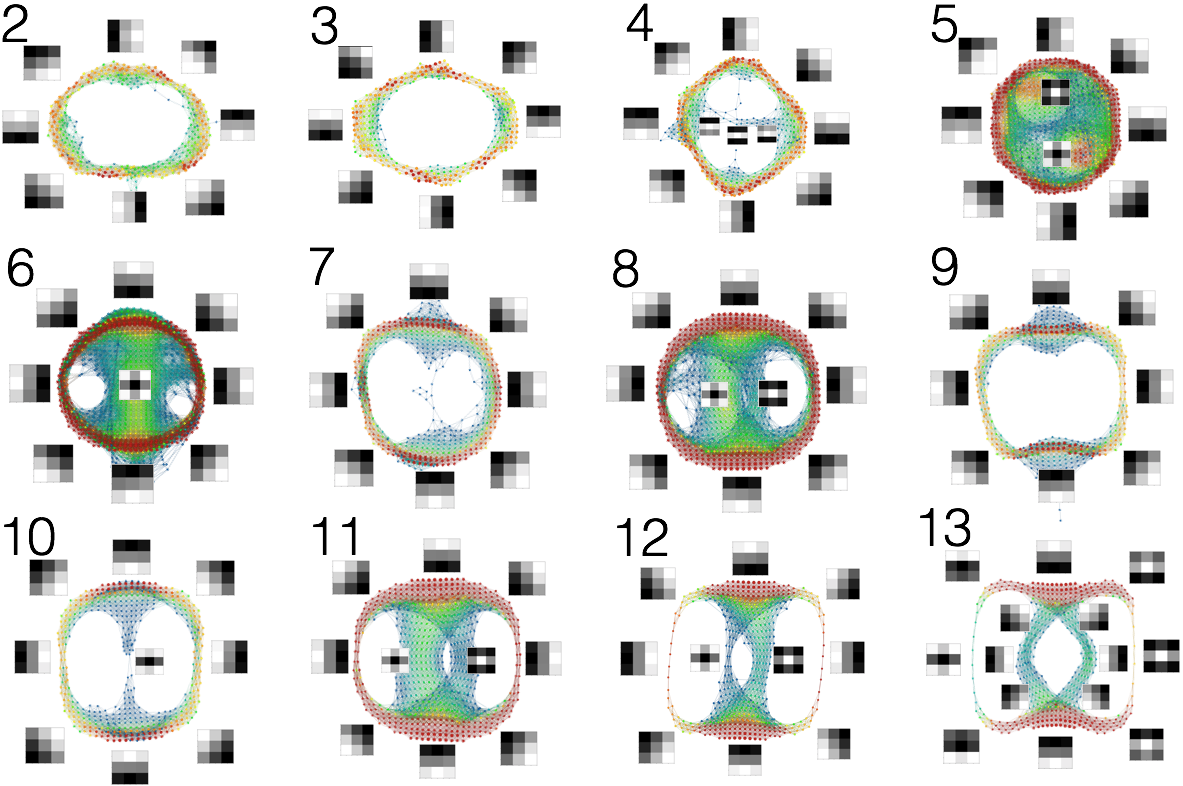}
\end{center}
  \caption{Mapper applied to the convolutional spatial filters of VGG16}
 \label{fig:vgg16}
\end{figure*}

\begin{figure}[h]
\begin{center}
\includegraphics[height=3.3cm]{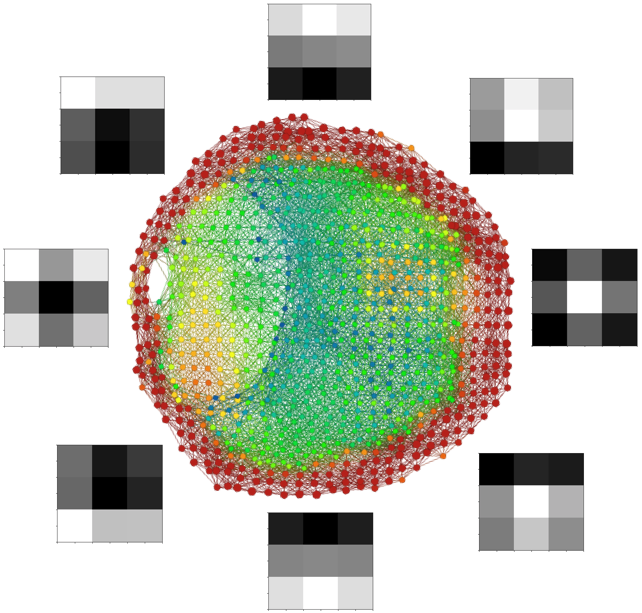}
\end{center}
 \caption{$Mapper(30,3)$ of $\rho(100, 0.3)$ of the fifth convolutional layer in VGG19}
  \label{fig:vgg19}
  \vspace{-0.5cm}
\end{figure}

\subsection{ImageNet and VGG}



We look at the spatial filters of a single pre-trained network VGG16 [4] trained on ImageNet. VGG16 contains 13 convolutional layers. The first layer only has $3\times 64=192$ spatial filters which proved too few to locate a significant topological structure using Mapper or Persistence. However, subsequent layers have many more spatial filters. In Figure \ref{fig:vgg16} we include the Mapper output of the 12 convolutional layers following the first layer. For each layer we use $Mapper(30,3)$ and for layer 3-13 we use $ \rho(100,0.3) $ while for layer 2 we use $ \rho(100,0.4) $.

In all but the last layer (layer 13) we find the primary circle as the dominant structure. We also find some patches that have no counterpart in the Klein bottle model in [2], notably in layers 5,6,8,11,12,and 13.  Note that they appear in the higher layers and may reflect things detected in higher layers in the human visual pathway. We also look at a pre-trained network VGG19 [4] where we find other dominant structures at certain layers, for example already at layer 5 in VGG19 we find the dominant circle in layer 13 of VGG16, see Figure \ref{fig:vgg19}. Also note that this circle closely resembles that found in Figure \ref{fig:cifar_1st_1layer}.

\section{Interpretation: A Measure of Generality}

In this section we demonstrate a connection between the simplicity of the topological structure of a network's learned weights and its ability to generalize to unseen data.

We look both at the network’s ability to generalize to a new dataset (Street House View Numbers, or SVHN [22]) and unseen data in form of held out test data. First we train a network of type $M (64, 32, 64)$ (Figure \ref{table:M}) on MNIST [5] under three different circumstances: (i) we fix the first convolutional layer to a perfect discretization of the primary circle (Figure \ref{fig:circles}), (ii) we fix the first convolutional layer to a random gaussian, and (iii) we train the network as in regular circumstances with nothing fixed. We train for 40,000 batch iterations (to test accuracies of about 99\%) and then evaluate all the three networks on SVHN (we train and evaluate each network three times and take the average of the evaluation accuracies). We test on 26,032 images of SVHN that we rescale to 28x28 and grayscale. We get the following test accuracies: (i) 28 \%, (ii) 12 \%, and (iii) 11 \%. This suggests that enforcing an idealized version of the topological structure found in the data helps improve the ability to generalize across different data sets. 
SVHN is a more diverse dataset that contains a greater variety of fonts and styles, including digits that look handwritten. Indeed, training a network on SVHN (to test accuracy of about 85\%) and then evaluating it on MNIST (50,000 images) we achieve a test accuracy of 54\%. Therefore, in line with the hypothesis that a simpler topology implies better generalization capabilities, we should expect the network to learn a simpler structure when trained on SVHN than on MNIST. We find that the first layer weights of the network trained on SVHN learn a primary circle as when trained on MNIST (Figure \ref{fig:mnist_1st_2layer}), only that the primary circle is stronger in the SVHN case: the lifetime (birth time subtracted from death time) of the most persistent (greatest lifetime) 1-homology is significantly greater when trained for 40,000 batch iterations on SVHN than on MNIST (1.27 versus 1.10), where we used filtration $\rho(100, 0.1)$ as defined in section 4. 

\begin{figure}[h]
\begin{center}
\includegraphics[width=0.486\linewidth]{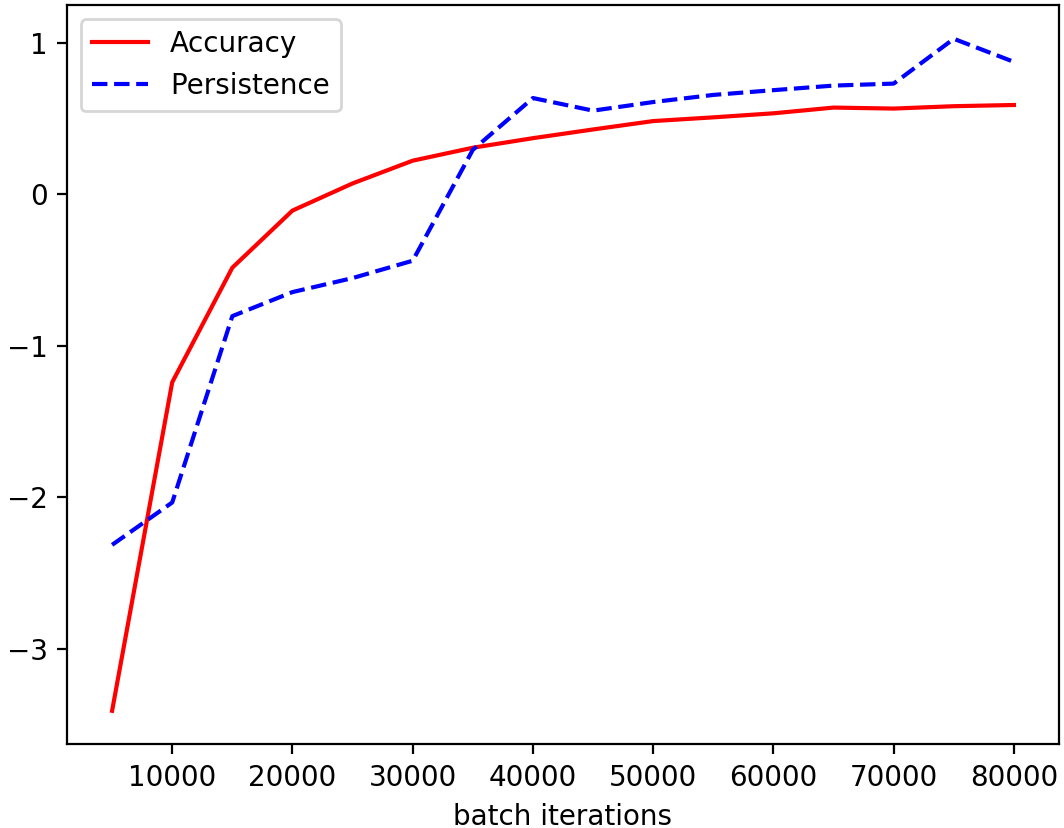}
\includegraphics[width=0.496\linewidth]{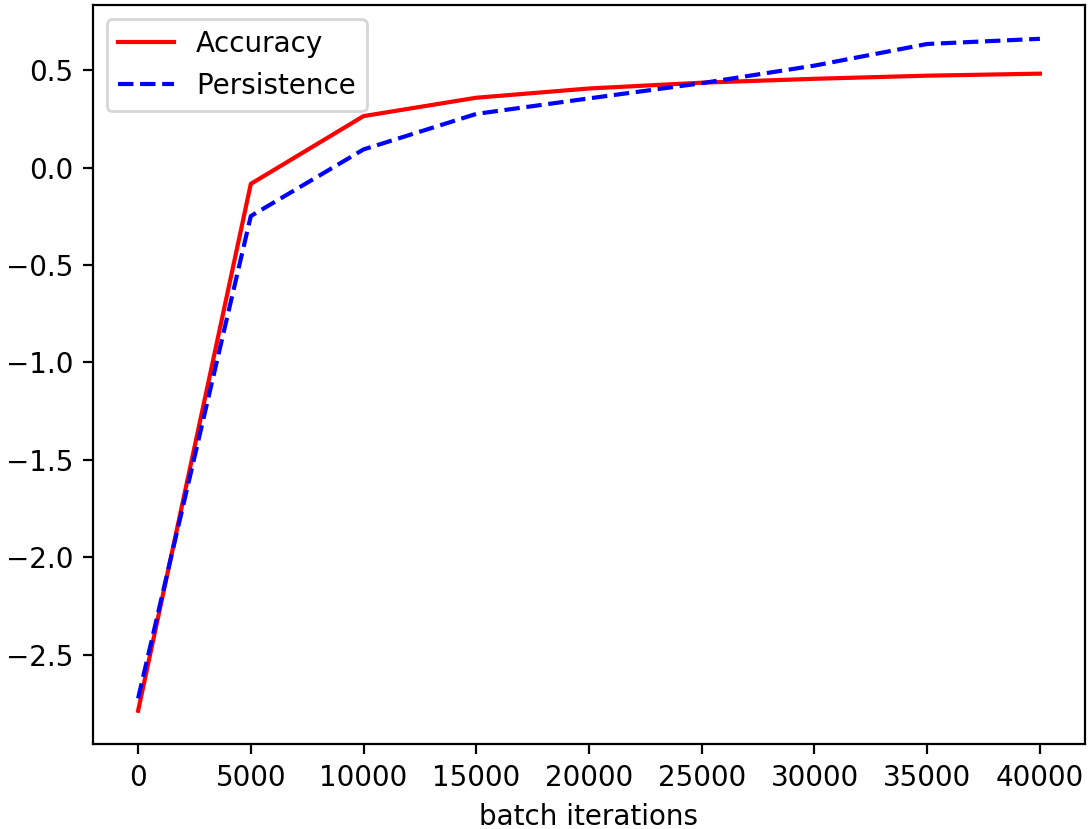} 
\end{center}
 \vspace{-0.20cm}
  \caption{Test Acc. and Persistence. Left: MNIST, right: SVHN }
  \label{fig:MNISTgen}
\end{figure}

Next, we train networks of type $M (64, 32, 64)$ on MNIST and SVHN and look at the correlation between the lifetime of the most persistent 1-homology of the spatial filters at filtration $\rho(100, 0.1)$ and the test-accuracy within the same domain (i.e. held out test data). We mean-center and normalize the data and plot the test accuracy and persistence versus number of batch iterations (Figure \ref{fig:MNISTgen}).

\begin{figure}[h]
\begin{center}
\includegraphics[height=5cm]{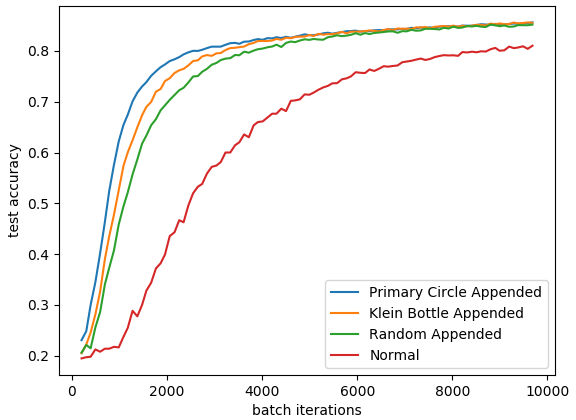} 
\end{center}
\vspace{-0.20cm}
  \caption{Network with appended information trained on SVHN. Test accuracy vs. batch iterations}
  \label{fig:SVHNappend}
  \vspace{-0.35cm}

\end{figure}

These results indicate a connection between test-accuracy and the lifetime of the most persistent 1-dim homology of the first-layer spatial filters, i.e. the 'topological simplicity.'

Lastly, we investigate the effect of appending idealized weight features found in our topological analysis to the raw input pixel values. To this end, we preprocessed each input image with a set of fixed $3\times3$ weights whose inner product with each $3\times3$ patch of the input image was appended to the central pixel value of the patch. We used three different sets of preprocessing weights: (1) 64 weights from the idealized primary circle found in Figure \ref{fig:mnist_1st_2layer}, (2) 64 weights from the idealized extension to the three-circle structure found in Figure \ref{fig:cifar_1st_2layer_c3}, i.e. the Klein bottle as per [2], and (3) a random gaussian. In Figure \ref{fig:SVHNappend} we plot the test accuracy versus the number of batch iterations for these three sets as well as for a "Normal" network without any appended preprocessed features, for networks trained on SVHN. Networks trained on MNIST under the same circumstances showed the same trend but the differences were smaller. In accordance with our findings of their relative strength in the weights of networks trained on MNIST or SVHN, the idealized primary circle provides the greatest improvements, followed by the Klein bottle. These both do better than a set of random gaussians. We found substantial improvement in the training time for both MNIST and SVHN when using the additional primary circle features. A factor of 2 speed up was realized for MNIST, and a factor of 3.5 for SVHN. MNIST is a much cleaner and therefore easier data set, and we suspect that the speed up will in general be larger for more complex data sets.

\section{Discussion}

We have demonstrated that topological modeling can be used as an effective tool to obtain understanding of the functioning of CNN's. Many results we found about the topological spaces of the trained weights were unexpected and non-trivial, and went beyond those of the motivating paper [2]. We have shown that the spaces of spatial filters learn simple global structures. This is true not only for the first layer, but occurs at least up to layers at depth 13. We also demonstrated the change of the simple structures over the course of training. We provided an interpretation of the topological structures we found and how they might be used. We showed that topological information can increase a network's ability to generalize to unseen data, be indicative of the generality of the dataset on which the network was trained, and it can improve and speed up the training of networks. We also showed a measure of the strength (or simplicity) of a topological feature and how it correlates with test accuracy on unseen test data. This lays the foundations for future work that further demonstrates and investigates the connection between the existence of simple topological models of the learned weight spaces on the one hand and the ability to generalize across data sets on the other. We see both how topological information may serve as a measure of generality as well as a potential regularizer.

\section{Acknowledgements}

We thank Primoz Skraba for his help and useful input. This work was supported by Altor Equity Partners AB through Unbox AI (unboxai.org).

\section*{References}
{\small
\bibliographystyle{ieee}

[1] Zeiler, Matthew D. \& Fergus, Rob. (2014) Visualizing and Understanding Convolutional Networks. {\it Computer Vision –ECCV 2014}, pp.\ 818--833. Springer International Publishing
  

[2] Carlsson, Gunnar and Ishkhanov, Tigran and de Silva, Vin and Zomorodian, Afra. (2008) On the Local Behavior of Spaces of Natural Images. {\it International Journal of Computer Vision} {\bf 76}(1)1-12

[3] Carlsson, Gunnar. (2009) Topology and Data. {\it Bulletin (New Series) of the American Mathematical Society} {\bf 46}(2) 255–308


[4] Karen Simonyan and Andrew Zisserman. (2014) Very Deep Convolutional Networks for Large-Scale Image Recognition. {\it CoRR } 1409.1556 

[5] Y. LeCun. The MNIST database of handwritten
digits. http://yann.lecun.com/exdb/mnist. 

[6] A. Krizhevsky. (2009) Learning multiple layers of features from
tiny images. Technical report, University of Toronto

[7] J. Deng, W. Dong, R. Socher, L.-J. Li, K. Li, and L. FeiFei. (2009)
Imagenet: A large-scale hierarchical image database. In
{\it IEEE Conference on Computer Vision and Pattern Recognition. }


[9] G. Carlsson. (2014) Topological pattern recognition for point cloud data. {\it Acta Numerica} {\bf 23} 289-368  

[10] G. Singh, F. Memoli, and G. Carlsson. (2007) Topological methods for the analysis of high dimensional data sets and 3D object recognition. {\it Point Based Graphics}, Prague

[11] M. Carriere and S. Oudot. (2015) Structure and stability of the 1-Dimensional Mapper. arXiv:1511.05823.

[12] Ayasdi, TDA and machine learning (2016) https://www.ayasdi.com/wp-content/uploads/\_downloads/wp-tda-and-machine-learning.pdf


[14] H. Sexton, J. Kloke. (2015) Systems and methods for capture of relationships within information. U.S. Patent. 14/639,954. Filed Mar. 5, 2015

[15] Tierny, Julien. (2017) Topological data analysis for scientific visualization. Mathematics and Visualization. {\em Springer, Cham}. ISBN: 978-3-319-71506-3; 978-3-319-71507-062-07

[16] Topological and statistical methods for complex data. Tackling large-scale, high-dimensional, and multivariate data spaces. Papers from the Workshop on the Analysis of Largescale, High-Dimensional, and Multi-Dimensional and Multi-Variate Data Using Topology and Statistics held in Le Barp, June 12–14, (2013). Edited by Janine Bennett, Fabien Vivodtzev and Valerio Pascucci. Mathematics and Visualization. Springer, Heidelberg, (2015). ISBN: 978-3-662-44899-1; 978-3-662-44900-4 94-06

[17] Karen Simonyan, Andrea Vedaldi, Andrew Zisserman. (2014) Deep Inside Convolutional Networks: Visualising Image Classification Models and Saliency Maps. arXiv:1312.6034  

[18] Anh Nguyen, Jason Yosinski, Jeff Clune. (2016) Multifaceted Feature Visualization: Uncovering the Different Types of Features Learned By Each Neuron in Deep Neural Networks. arXiv:1602.03616v2

[19] Anh Nguyen, Alexey Dosovitskiy, Jason Yosinski, Thomas Brox, Jeff Clune. (2016) Synthesizing the preferred inputs for neurons in neural networks via deep generator networks. arXiv:1605.09304

[20] A. Mahendran and A. Vedaldi. (2015) Understanding deep image representations by inverting them. {\em IEEE Conference on Computer Vision and Pattern Recognition (CVPR)}, Boston, MA, pp. 5188-5196

[21] Abhishek Sinha, Mausoom Sarkar, Aahitagni Mukherjee, Balaji Krishnamurthy. (2017) Introspection: Accelerating Neural Network Training By Learning Weight Evolution. 	arXiv:1704.04959 

[22] Yuval Netzer, Tao Wang, Adam Coates, Alessandro Bissacco, Bo Wu, Andrew Y. Ng. (2011) Reading Digits in Natural Images with Unsupervised Feature Learning NIPS Workshop on Deep Learning and Unsupervised Feature Learning

}

\end{document}